\documentclass[letterpaper]{article}
\usepackage{iccc}

\usepackage{times}
\usepackage{algorithm}
\usepackage{algpseudocode}
\usepackage{helvet}
\usepackage{graphicx}
\usepackage{courier}
\title{Prompt to GPT-3: Step-by-Step Thinking Instructions for Humor Generation}
\author{Yuetian Chen, Bowen Shi, Mei Si\\
Rensselaer Polytechnic Institute\\
110 8th street, 
Troy, New York 12180\\
\{cheny63, shib5, sim\}@rpi.edu\\
}
\begin{document} 
\maketitle
\begin{abstract}
\begin{quote}
Artificial intelligence has made significant progress in natural language processing, with models like GPT-3 demonstrating impressive capabilities. However, these models still have limitations when it comes to complex tasks that require an understanding of the user, such as mastering human comedy writing strategies. This paper explores humor generation using GPT-3 by modeling human comedy writing theory and leveraging step-by-step thinking instructions. In addition, we explore the role of cognitive distance in creating humor. 
\end{quote}
\end{abstract}

\section{Introduction}
Artificial Intelligence (AI) advancements have made significant progress in various fields, particularly natural language processing. In particular, GPT-3 has demonstrated impressive capabilities in many language generation tasks  \cite{brown2020language}. This paper explores using GPT3 to generate jokes in an explainable and controlled way. As the demand for personalized and engaging content in the entertainment industry grows, there is an increasing need for AI models that can produce humorous content that resonates with human audiences. Therefore, it is crucial to develop AI models that can generate humor in a way that is compatible with human preferences.

\begin{figure*}[h]
    \centering
    \includegraphics[width=0.8\linewidth]{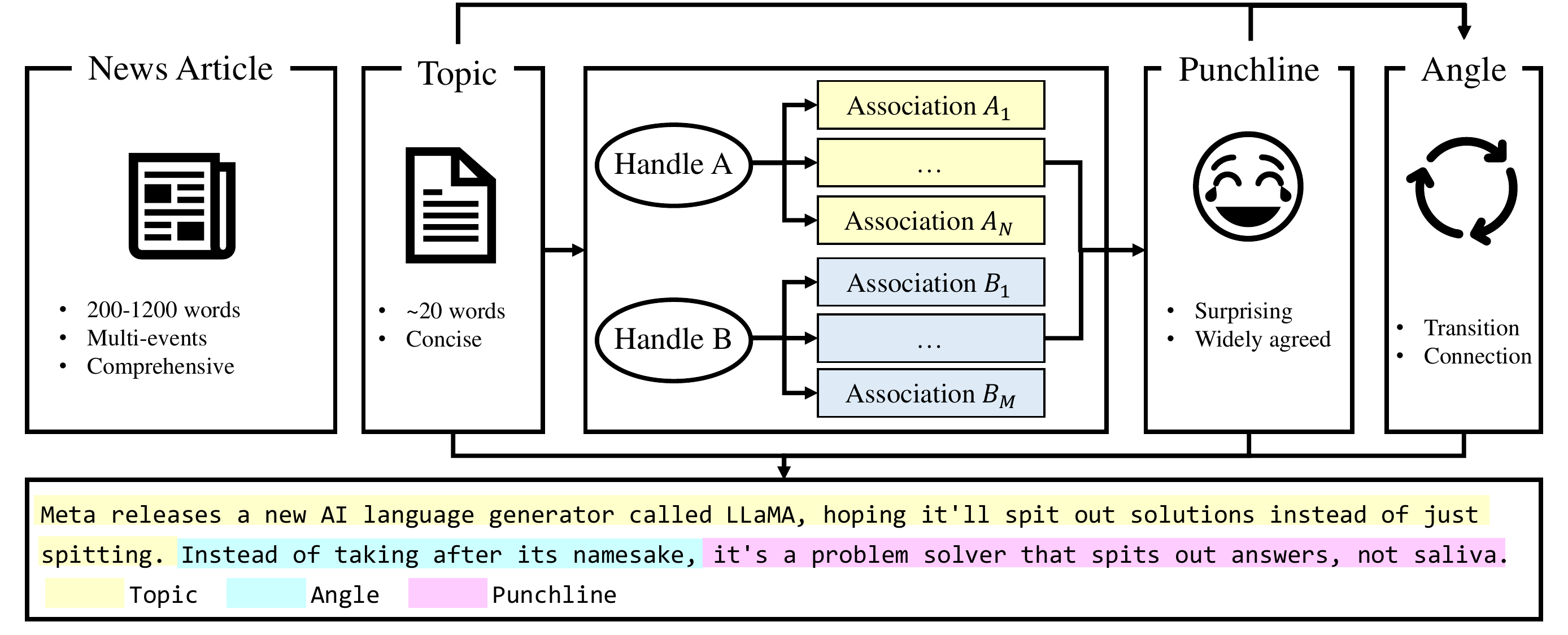}
    \caption{An overview of Joe Toplyn's theory in creating monologue jokes}
    \label{fig:pipe}
\end{figure*}


In the field of humor generation, there are two main approaches: template-based and neural network-based methods. Template-based methods, such as those used by He et al.\cite{he2019pun} and Castro et al.\cite{castro2016joke}, rely on predefined structures that are filled with appropriate words or phrases to create jokes. While these methods are simple and easy to implement, they are limited by the predefined templates and the availability of suitable words or phrases.

On the other hand, neural network-based methods, such as those used by Zhang et al.\cite{zhang2020let}, and Akbar et al.\cite{akbar2021deep}, utilize machine learning techniques. Zhang et al. used a neural network model to generate humorous captions for images by incorporating relevant knowledge from external sources. Akbar et al. fine-tuned a large pre-trained language model (GPT-2) on a joke dataset to generate jokes. These methods have the advantage of being able to learn from data and generate more diverse and original humor. However, the generation process is opaque to human users, and there is no way for people to understand how the model came up with the joke or instruct the model to generate jokes in a particular way. This may lead to the production of inappropriate or offensive humor. 


This paper investigates how to enhance GPT-3's ability to generate humor in an explainable fashion by integrating Joe Toplyn -- a famous late-night show writer's comedy writing theory \cite{toplyn2014comedy} through step-by-step thinking instructions to GPT3. We also explore the role cognitive distance plays in creating a humor effect. 

\section{Toplyn's Theory for Writing Jokes}
Toplyn is a renowned comedy writer who has worked on shows like ``Late Night with David Letterman" and ``The Tonight Show with Jay Leno." He outlines his process for creating humorous content in his book ``Comedy Writing for Late-Night TV: How to Write Monologue Jokes, Desk Pieces, Sketches, Parodies, Audience Pieces, Remotes, and Other Short-Form Comedy."~\cite{toplyn2014comedy} According to Toplyn, there are four steps to crafting a joke:

\begin{enumerate}
\item \textbf{Create a topic sentence based on a news item:} 
A news article is typically used as a starting point. The first step involves generating a topic sentence highlighting the news item's main point. The aim is to create a sentence that is engaging and appropriate for humorous commentary, while still being factually accurate and not inherently funny.


\item \textbf{Identify handles and associations:} Handles refer to interesting or peculiar words or phrases found within the topic sentence, and Toplyn usually identifies two handles for each joke. After identifying handles, in the next step, associations are created. Associations are the concepts or ideas related to each handle. Toplyn generates a list of associations for each handle, which are used in the subsequent step to develop a punchline.

\item \textbf{Develop a punchline by combining associations:} A punchline connects one association from the first handle's list with another from the second handle's list. This combination should be perceived as true by most people and evoke a negative emotion towards the first major entity in the topic. The negative emotion is essential for generating humor in the monologue joke.

\item \textbf{Connect the topic and punchline with an angle:} The angle is a sentence or phrase that smoothly transitions the audience from the topic to the punchline. It ensures that the joke has a coherent structure and flows naturally from beginning to end.
\end{enumerate}




The following is an example of a joke written by Joe: ``At Lax, customs inspectors seized a live shipment of 67 Giant African snails. Instead of destroying them, officials gave them jobs at the DMV." Joe Toplyn created an automated joke generation system based on his theory by prompting GPT-3~\cite{toplyn2023witscript}. This work is in the same spirit. We built upon Toplyn's approach by introducing a new factor - the cognitive distance between associations for creating punchlines - to generate more effective jokes. This involves identifying seemingly unrelated associations to craft punchlines, resulting in better jokes.

\section{Implementation and Examples}


In our method, we adopt step-by-step thinking instructions that combine a chain of thought reasoning~\cite{wei2023chainofthought} with Toplyn’s theory to guide GPT-3 in generating jokes. This approach includes a set of well-produced prompts for each stage of the process, allowing the model to focus on individual aspects of the joke sequentially. The chain of thought reasoning and iterative report generation during each prompt help to ensure a coherent and consistent joke creation process. The examples of intermediate results and final jokes are provided in Table~\ref{table:joke_example1}.


\subsubsection{Creating the Topic Sentence from News Article}


The instruction we used is as follows. It defines what a topic sentence is with examples:

\begin{itemize}
    \item [] \textit{``We will generate Monologue joke topics for a late-night TV show by crafting succinct sentences based on actual news items. A Monologue joke comprises three parts, and our objective is to produce the first part - the topic. The topic should be founded on a real news event that grabs people's attention and enables amusing commentary. It need not be intentionally funny but must be factually accurate. For instance, `Carl's Jr. is selling a foot-long burger' or `Bernie Madoff's underpants were sold at an auction' are suitable topics. During our conversation, I will provide a news article, and you will create a single sentence that fulfills these criteria. If you believe the news article is inappropriate for Monologue jokes, please inform me."
}
\end{itemize}
Based on our experience, medium-sized articles (500 to 800 words) work best for generating jokes. This is because they provide enough context to write a relevant and engaging joke, while not being so long that they become overwhelming. Articles that are too short may lack enough context for a joke, while articles that are too long may contain extraneous information that dilutes the comedic focus. 




\subsubsection{Identifying Handles and Associations}


Handles are potential humor generators and can be people, places, things, or actions. In the prompt, we provide examples to ensure accuracy in identifying handles for a richer punchline. 

\begin{itemize}
    \item [] \textit{``To further develop this monologue joke for a late-night TV show based on the provided topic sentence, we need to create the punchline, which is the surprise element at the end of the joke. To do this, follow these steps:
    \begin{itemize}
        \item Determine two handles in the topic, which are interesting words or phrases. Handles can include people, places, things, or actions.
        \item Brainstorm a list of associations for each handle and then create two separate lists of associations related to each handle.
    \end{itemize}   
    Now, based on the topic you provided, identify handles and associations in the format given by the example."}
\end{itemize}

\begin{table*}[h]
    \centering
    \begin{tabular}{|p{5cm}|p{10cm}|}
    \hline
    \textbf{Component}                & \textbf{Description} \\ \hline
    \textit{Topic}                    & Microsoft introduces a new AI-powered Copilot for their 365 apps, making Clippy's ghost proud. \\ \hline
    \textit{Handles}                  & AI-powered Copilot, Clippy's ghost \\ \hline
    \textit{Associations for \textbf{AI-powered Copilot}} & Artificial intelligence, \textbf{Clippy 2.0}, Microsoft 365, Productivity, GPT-4, Virtual assistant, \textbf{Automated tasks}, Office apps, Innovative technology \\ \hline
    \textit{Associations for \textbf{Clippy's ghost}} & Nostalgia, Old technology, \textbf{Revolutionary technology}, Paperclip, \textbf{Annoying assistant}, Pop-up help, Microsoft Office, 90s tech, Failed innovation \\ \hline
    \textit{Punchline (\textbf{Negative})}                &  \textbf{Automated tasks} + \textbf{Annoying assistant}: Now, it can automatically annoy you with its help. \\ \hline
    \textit{Punchline (\textbf{Positive})}                & \textbf{Clippy 2.0}+\textbf{Revolutionary technology}: Clippy's cool cousin has arrived! \\ \hline
    \textit{Angle (\textbf{Negative})}                    & In the spirit of Clippy, \\ \hline
     \textit{Angle (\textbf{Positive})}                    & It turning your office into a futuristic workspace, with one chatbot at a time - \\ \hline
    \textit{Final Joke (\textbf{Negative})}               & Microsoft introduces a new AI-powered Copilot for their 365 apps, making Clippy's ghost proud. In the spirit of Clippy, now it can automatically annoy you with its help. \\ \hline
    \textit{Final Joke (\textbf{Postive})}               & Microsoft introduces a new AI-powered Copilot for their 365 apps, making Clippy's ghost proud. It turning your office into a futuristic workspace, with one chatbot at a time - Clippy's cool cousin has arrived! \\ \hline
    \end{tabular}
    \caption{Example of joke generation using the proposed method with positive and negative emotion instruction}
    \label{table:joke_example1}
\end{table*}

\subsubsection{Developing the Punchline}


The prompt provided below emphasizes the significance of eliciting a negative feeling towards the first major entity in the topic sentence. This is important because it creates a target for the joke, which allows the audience to laugh at the joke more easily.

\begin{itemize}
    \item [] \textit{``Pair an association from one list with an association from the other list. Choose a combination that most people would perceive as true to create the punchline. It is important to evoke a \textbf{negative} emotion towards the first major entity in the topic for the monologue joke to be humorous.
    Now, based on the association lists you provided, provide the punchline as shown in the example.''}
\end{itemize}

Our study demonstrates that the inclusion of negative emotion in a joke plays a crucial role in its humor. Table~\ref{table:joke_example1} provides two instances of jokes addressing the same topic but with different sentiment prompts. When the sentiment in the prompt is changed from ``Negative" to ``Positive," the joke may lose its humor. This can be attributed to the fact that the joke prompted with the ``Negative" sentiment keyword contains more negative emotion, resulting in greater contrast between the subjects involved.

\subsubsection{Connecting the Topic and Punchline with an Angle}


In the final step, we guide GPT-3 to create a smooth link between the topic sentence and the punchline. Due to the lengthiness of the complete prompt, it is not possible to provide it here. Essentially, the previously generated content is used as input to provide context. Then, the instruction for forming the punchline is then given as follows:
\begin{itemize}
    \item [] \textit{``Next, craft an angle to smoothly transition the audience from the topic to the punchline.''}
\end{itemize}

The example included in the prompt provides guidance on how to design an effective angle. Specifically, given the topic and the punchline as follows:
\\\\
\texttt{\textbf{Topic: }The FBI is warning people about the dangers of charging devices in public areas, especially hotel lobbies.}
\texttt{\textbf{Punchline: }Your device might end up with more than just bed bugs.}
\\\\
One of the angles could be:
\\\\
\texttt{\dots If you are using a USB port in a hotel lobby, \dots}
\\\\
Again, the example within the prompt reinforces the model's understanding of how to effectively weave together the various elements of the joke, transforming the topic, punchline, and associations into a cohesive and entertaining monologue joke. By combining the output generated in each stages, we can get a final report shown in Algorithm \ref{alg:report}.


\begin{algorithm}[h]
    \caption{Report for the new "1080p Premium" feature on YouTube}
    \label{alg:report}
    \begin{algorithmic}[1]
        \State \textbf{Topic:} ``YouTube experiments with a new ``1080p Premium'' option, offering higher-quality video for Premium subscribers.''
        
        \State \textbf{Associations for ``1080p Premium'':}
        \begin{itemize}
            \item ``Premium subscribers''
            \item ``YouTube revenue''
            \item ``Exclusive content''
            \item ``Higher price''
            \item \textbf{``Upselling''}
            \item ``Better experience''
        \end{itemize}
        
        \State \textbf{Associations for ``higher-quality video'':}
        \begin{itemize}
            \item ``HD resolution''
            \item ``Better image quality''
            \item ``More pixels''
            \item ``Higher bitrate''
            \item \textbf{``Sharper image''}
            \item ``More data''
            \item ``Cinematic experience''
        \end{itemize}
        
        \State \textbf{Punchline:} ``Upselling their way to a sharper disappointment.''
        \State \textbf{Angle:} ``Now, viewers can pay more to witness their favorite cat videos in stunning clarity.''
        
        \State \textbf{Summary:} ``YouTube experiments with a new ``1080p Premium'' option, offering higher-quality video for Premium subscribers. Now, viewers can pay more to witness their favorite cat videos in stunning clarity – upselling their way to a sharper disappointment.''
    \end{algorithmic}
\end{algorithm}

\subsection{Example Outputs}
In this section, we present additional examples of jokes generated using our proposed method\footnote{GitHub repository: https://github.com/Stry233/Joe-Toplyn-s-Monologue-Joke-generation-using-ChatGPT}. 


The first example's topic is Nintendo's decision to skip E3 2023, leaving gamers curious about what they have in store. The punchline creatively links ``surprise announcements" and ``childhood memories," by replacing the former with the game "hide-and-seek." The resulting joke is: ``Nintendo decides to skip E3 2023, leaving gamers wondering what's up their sleeve. Turns out, their new game plan is to play hide-and-seek with our childhood memories."

The second example involves Meta's new AI language generator, LLaMA. The topic centers on Meta's hope that LLaMA will generate solutions instead of just spitting. By connecting the associations of ``AI language generator" and ``LLaMA," the punchline is crafted: ``Meta releases a new AI language generator called LLaMA, hoping it'll spit out solutions instead of just spitting. Instead of taking after its namesake, it's a problem solver that spits out answers, not saliva."

The third example, as shown in Table~\ref{table:joke_example5}, makes connections among three associations. It involves a Japanese spacecraft that's set to make a historic moon landing. Its cargo includes the UAE's rover and a lunar robot made by a Japanese toy maker. By combining the associations of ``historic moon landing," ``UAE's rover," and ``Japanese toy maker," we arrived at the punchline: ``As the Japanese spacecraft lands on the moon, carrying the UAE's rover and a lunar robot from a Japanese toy maker, Neil Armstrong's famous quote gets a cosmic update: `One giant leap for UAE's rover, one small step for anime-kind'."
\begin{table*}[h]
    \centering
    \begin{tabular}{|p{5cm}|p{10cm}|}
    \hline
    \textbf{Component}                & \textbf{Description} \\ \hline
    \textit{Topic}                    & A Japanese spacecraft is attempting a historic moon landing, delivering the UAE's rover and a lunar robot from a Japanese toy maker. \\ \hline
    \textit{Handles}                  & historic moon landing, UAE's rover, Japanese toy maker \\ \hline
    \textit{Associations for \textbf{historic moon landing}} & \textbf{Neil Armstrong}, Moonwalk, One small step for man, Apollo, Space race, Lunar surface \\ \hline
    \textit{Associations for \textbf{UAE's rover}} & \textbf{United Arab Emirates}, Space exploration, Desert, Arabian Nights, Sand dunes \\ \hline
    \textit{Associations for \textbf{Japanese toy maker}} & \textbf{Anime}, Gundam, Hello Kitty, Action figures, Remote control toys, Collector's items \\ \hline
    \textit{Punchline}                & Neil Armstrong's famous quote is updated: ``One giant leap for UAE's rover, one small step for anime-kind. \\ \hline
    \textit{Angle}                    & In a cosmic twist, \\ \hline
    \textit{Final Joke}               & A Japanese spacecraft is attempting a historic moon landing, delivering the UAE's rover and a lunar robot from a Japanese toy maker. In a cosmic twist, Neil Armstrong's famous quote is updated: ``One giant leap for UAE's rover, one small step for anime-kind." \\ \hline
    \end{tabular}
    \caption{Example of joke generation using the proposed method}
    \label{table:joke_example5}
\end{table*}

\subsection{Create Punchlines using Unrelated Concepts}
Although not mentioned in Toplyn's theory, we found that the selection of associations is a critical factor in creating engaging and humorous jokes. Selecting associations that are less obviously related can lead to more unexpected and intriguing punchlines. These types of punchlines often have a stronger comedic effect, which enhances the humor of the joke. In contrast, if we were to select associations that are more closely related, the resulting punchline may not be surprising enough to provoke laughter or amusement. In such cases, the joke might feel predictable or mundane instead. 




For example, let's say we have two handles - ``Space Travel" and ``Fast Food." We retrieve associations for each handle as follows:
\\\\
    \texttt{
       Space Travel: Mars, freeze-dried meal, astronaut\\
       Fast Food: burger, fries, drive-thru
    }
\\\\
When picking two associations for creating the punchline, we find that the pair \texttt{Mars} and \texttt{burger} are mostly irrelevant to each other. Using this pair, we can create a punchline that more likely surprises the audience and creates humor. For example:
\\\\
    \texttt{Why did the astronaut bring a \textbf{burger} to \textbf{Mars}? Because he heard it was a great place for a `space'cial!}
\\\\
Alternatively, if we choose to use \texttt{freeze-dried meal} and \texttt{burger} as a word-pair which has a relatively higher semantic relevance score.
\\\\
    \texttt{Why did the astronaut prefer a \textbf{freeze-dried meal} over a \textbf{burger}? Because it's easier to pack for space travel!}
\\\\
This punchline lacks humor because the connection between the two concepts is easily understandable and lacks an element of surprise. The audience can quickly identify the relationship between a freeze-dried meal for astronauts and a burger, which reduces the effectiveness of the punchline in creating laughter.

\section{Discussion and Future Work}

Our work demonstrates that it is possible to use step-by-step instructions to guide GPT-3 in following human comedy writing theory, resulting in the generation of configurable and explainable jokes. The jokes are configurable because, although the process can be fully automated, humans can intervene and modify the intermediate results in the joke creation process. This advancement opens up new possibilities for AI-driven humor and entertainment and contributes to the growing body of research in the field of AI and humor studies.

The next step in our research involves incorporating additional comedy writing theories and techniques. This includes exploring various types of humor like irony, satire, and sarcasm, as well as more advanced techniques such as wordplay, puns, and misdirection. We also plan to examine the potential of integrating user feedback into the system, where users can rate and provide feedback on the generated jokes, which will be used by reinforcement learning process to improve the overall quality and effectiveness of the joke generation process. 

We also plan to further study the relationship between the level of difficulty in creating connections between associations and the resulting perceived humor. Specifically, we want to investigate whether the level of cognitive effort required to make sense of seemingly unrelated associations enhances or detracts from the overall comedic effect. This could provide valuable insights into how to optimize the joke generation process for maximum comedic impact.

Finally, we want to personalize the system by considering users' humor preferences. By gathering data on users' preferred humor styles, comedians, and amusing joke types, we can train the system to create jokes that match their sense of humor. Additionally, we can try creating jokes for specific events or holidays to enhance their relevance and humor value.



\bibliographystyle{iccc}
\bibliography{iccc}

\begin{thebibliography}{}

\bibitem[\protect\citeauthoryear{Akbar \bgroup et al.\egroup
  }{2021}]{akbar2021deep}
Akbar, N.~A.; Darmayanti, I.; Fati, S.~M.; and Muneer, A.
\newblock 2021.
\newblock Deep learning of a pre-trained language model's joke classifier using
  gpt-2.
\newblock {\em Journal of Hunan University Natural Sciences} 48(8).

\bibitem[\protect\citeauthoryear{Brown \bgroup et al.\egroup
  }{2020}]{brown2020language}
Brown, T.~B.; Mann, B.; Ryder, N.; Subbiah, M.; Kaplan, J.; Dhariwal, P.;
  Neelakantan, A.; Shyam, P.; Sastry, G.; Askell, A.; Agarwal, S.;
  Herbert-Voss, A.; Krueger, G.; Henighan, T.; Child, R.; Ramesh, A.; Ziegler,
  D.~M.; Wu, J.; Winter, C.; Hesse, C.; Chen, M.; Sigler, E.; Litwin, M.; Gray,
  S.; Chess, B.; Clark, J.; Berner, C.; McCandlish, S.; Radford, A.; Sutskever,
  I.; and Amodei, D.
\newblock 2020.
\newblock Language models are few-shot learners.

\bibitem[\protect\citeauthoryear{Castro \bgroup et al.\egroup
  }{2016}]{castro2016joke}
Castro, S.; Cubero, M.; Garat, D.; and Moncecchi, G.
\newblock 2016.
\newblock Is this a joke? detecting humor in spanish tweets.
\newblock In {\em Advances in Artificial Intelligence-IBERAMIA 2016: 15th
  Ibero-American Conference on AI, San Jos{\'e}, Costa Rica, November 23-25,
  2016, Proceedings 15},  139--150.
\newblock Springer.

\bibitem[\protect\citeauthoryear{He, Peng, and Liang}{2019}]{he2019pun}
He, H.; Peng, N.; and Liang, P.
\newblock 2019.
\newblock Pun generation with surprise.
\newblock {\em arXiv preprint arXiv:1904.06828}.

\bibitem[\protect\citeauthoryear{Toplyn}{2014}]{toplyn2014comedy}
Toplyn, J.
\newblock 2014.
\newblock {\em Comedy Writing for Late-Night TV: How to Write Monologue Jokes,
  Desk Pieces, Sketches, Parodies, Audience Pieces, Remotes, and Other
  Short-Form Comedy}.

\bibitem[\protect\citeauthoryear{Toplyn}{2023}]{toplyn2023witscript}
Toplyn, J.
\newblock 2023.
\newblock Witscript 3: A hybrid ai system for improvising jokes in a
  conversation.
\newblock {\em arXiv preprint arXiv:2301.02695}.

\bibitem[\protect\citeauthoryear{Wei \bgroup et al.\egroup
  }{2023}]{wei2023chainofthought}
Wei, J.; Wang, X.; Schuurmans, D.; Bosma, M.; Ichter, B.; Xia, F.; Chi, E.; Le,
  Q.; and Zhou, D.
\newblock 2023.
\newblock Chain-of-thought prompting elicits reasoning in large language
  models.

\bibitem[\protect\citeauthoryear{Zhang \bgroup et al.\egroup
  }{2020}]{zhang2020let}
Zhang, H.; Liu, D.; Lv, J.; and Luo, C.
\newblock 2020.
\newblock Let's be humorous: Knowledge enhanced humor generation.
\newblock {\em arXiv preprint arXiv:2004.13317}.

\end{thebibliography}

\end{document}